\documentclass[
]{ceurart}

\usepackage{listings}
\usepackage{hyperref}
\usepackage{url}
\usepackage{multirow}
\usepackage{wrapfig}

\usepackage[utf8]{inputenc} 
\usepackage[T1]{fontenc}    
\usepackage{hyperref}       
\usepackage{url}            
\usepackage{booktabs}       
\usepackage{amsfonts}       
\usepackage{nicefrac}       
\usepackage{microtype}      
\usepackage{amsmath,amsthm,amssymb}
\usepackage{graphicx}
\usepackage{verbatim}
\usepackage{float}
\usepackage{color}
\usepackage{dirtytalk}
\usepackage{bbm}
\usepackage{enumitem}
\usepackage[ruled,linesnumbered]{algorithm2e}
\usepackage{xcolor}
\usepackage{multirow}
\usepackage{pgfplots}
\usepackage{threeparttable}
\usepackage{balance}  
\usepackage{tikz}
\usepackage{subfig}

\usepackage{graphicx}
\usepackage{xfrac}

\def \COMPLETE{}

\newtheorem{theorem}{Theorem}[section]

\newtheorem{definition}{Definition}

\newcommand{\mc}[1]{\mathcal{ #1}}
\newcommand{\mb}[1]{\mathbf{ #1}}

\newcommand{\supp}{\ifdefined\COMPLETE appendix \else supplementary \fi}

\newcommand\numberthis{\addtocounter{equation}{1}\tag{\theequation}}

\graphicspath{{figures/}}

\usepackage{amsmath,amsfonts,bm}

\def\eqref#1{equation~\ref{#1}}

\def\1{\bm{1}}

\DeclareMathAlphabet{\mathsfit}{\encodingdefault}{\sfdefault}{m}{sl}
\SetMathAlphabet{\mathsfit}{bold}{\encodingdefault}{\sfdefault}{bx}{n}

\DeclareMathOperator*{\argmax}{arg\,max}

\begin{document}

\copyrightyear{2022}
\copyrightclause{Copyright for this paper by its authors.
  Use permitted under Creative Commons License Attribution 4.0
  International (CC BY 4.0).}

\conference{In A. Martin, K. Hinkelmann, H.-G. Fill, A. Gerber, D. Lenat, R. Stolle, F. van Harmelen (Eds.),
Proceedings of the AAAI 2022 Spring Symposium on Machine Learning and Knowledge Engineering for Hybrid Intelligence (AAAI-MAKE 2022),
Stanford University, Palo Alto, California, USA, March 21–23, 2022.}

\title{Adjoined Networks: A Training Paradigm with Applications to Network Compression}

\author[1]{Utkarsh Nath}[%
email=unath@asu.edu,
]
\address[1]{School of Computing and Augmented Intelligence, Arizona State University, Tempe, AZ 85281, USA}

\author[2]{Shrinu Kushagra}[%
email=skushagr@uwaterloo.ca,
]
\address[2]{University of Waterloo, Waterloo, ON N2L 3G1, Canada}

\author[1]{Yingzhen Yang}[%
email=yingzhen.yang@asu.edu,
]

\begin{abstract}
Compressing deep neural networks while maintaining accuracy is important when we want to deploy large, powerful models in production and/or edge devices. One common technique used to achieve this goal is knowledge distillation. Typically, the output of a static pre-defined teacher (a large base network) is used as soft labels to train and transfer information to a student (or smaller) network. In this paper, we introduce \textit{Adjoined Networks}, or AN, a learning paradigm that trains both the original base network and the smaller compressed network together. In our training approach, the parameters of the smaller network are shared across both the base and the compressed networks. Using our training paradigm, we can simultaneously compress (the student network) and regularize (the teacher network) any architecture. In this paper, we focus on popular CNN-based architectures used for computer vision tasks. We conduct an extensive experimental evaluation of our training paradigm on various large-scale datasets. Using ResNet-50 as the base network, AN achieves 71.8\% top-1 accuracy with only 1.8M parameters and 1.6 GFLOPs on the ImageNet data-set. We further propose Differentiable Adjoined Networks (DANs), a training paradigm that augments AN by using neural architecture search to jointly learn both the width and the weights for each layer of the smaller network. DAN achieves ResNet-50 level accuracy on ImageNet with $3.8\times$ fewer parameters and $2.2\times$ fewer FLOPs.
\end{abstract}


\maketitle

\section{Introduction}
	
	\begin{figure}[!hbt]
        \begin{center}
            \includegraphics[scale=0.49, trim=10 30 0 60] {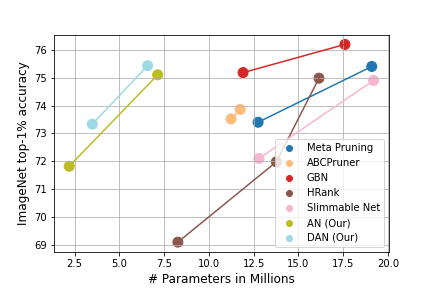}
        \end{center}

        \caption{Top-1 accuracy of various structured pruning methods (by compressing the ResNet-50 or ResNet-100 architecture) on the ImageNet dataset plotted against the number of parameters in the model. Our methods, Adjoined Networks (AN), and Differentiable Adjoined Networks (DANs) achieve similar accuracy as compared against current SOTA pruning methods but with fewer (in many cases 2x fewer) parameters.}
        \label{fig:imagenetComparison}
    \end{figure}

	Deep Neural Networks (DNNs) have achieved state-of-the-art performance on many tasks such as classification, object detection and image segmentation. However, the large number of parameters often required to achieve the performance makes it difficult to deploy them at the edge (like on mobile phones, IoT and embedded devices, etc). Unlike cloud servers, these edge devices are constrained in terms of memory, compute, and energy resources. A large network performs a lot of computations, consumes more energy, and is difficult to transport and update. A large network also has a high prediction time per image. This is a constraint when real-time inference is needed. Thus, compressing neural networks while maintaining accuracy and improving inference time has received significant attention in the last few years. Popular techniques for network compression include pruning and knowledge distillation.
	
	Pruning methods remove parameters (or weights) of overparameterized DNNs based on some pre-defined criteria. For example, \cite{han2015deep} removes weights whose absolute value is smaller than a threshold. While weight pruning methods are successful at reducing the number of parameters of the network, they often work by creating spares tensors that may require special hardware \cite{han2016eie} or special software \cite{park2017faster} to provide inference time speed-ups. These methods are also known as \textit{unstructured pruning} and has been extensively studied in \cite{han2015deep, zhu2017prune, gale2019state, kusupati2020soft, evci2021rigging}. To overcome this limitation, channel pruning \cite{liu2017learning} and filter pruning \cite{li2016pruning} techniques are used. These \textit{structured pruning} methods work by removing entire convolution channels or sometimes even filters based on some pre-defined criteria and can often provide significant improvement in inference times. In this paper, we show that our algorithm, Adjoined Networks or AN, achieves accuracy similar to the current state-of-the-art structured pruning methods but uses a significantly lower number of parameters and FLOPs (Fig \ref{fig:imagenetComparison}).
	
\begin{figure*}[t]
	\vspace{-10pt}
		\centering
		\includegraphics[scale=0.95, trim=0 0 0 10] {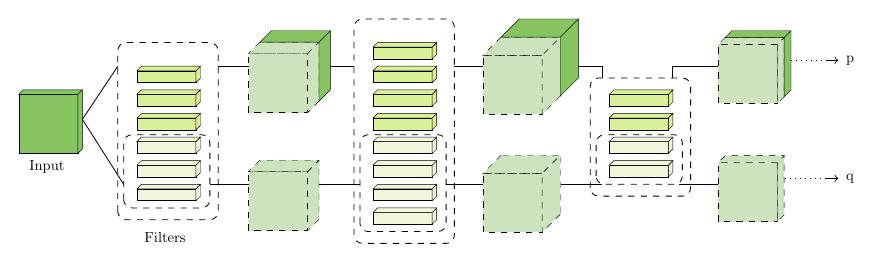}
		\vspace{-10pt}
		\caption{Training paradigm based on adjoined networks. The original and the compressed version of the network are trained together with the parameters of the smaller network shared across both. The network outputs two probability vectors $p$ (original network) and $q$ (smaller network).}
		\vspace{-10pt}
		\label{fig:adjoinedNetwork}
	\end{figure*}

    The AN training paradigm works as follows. A given input image $X$ is processed by two networks, the larger network (or the base network) and the smaller network (or the compressed network). The base network outputs a probability vector $p$ and the compressed network outputs a probability vector $q$. This setup is similar to the student-teacher training used in Knowledge Distillation \cite{hinton2015distilling} where the base network (or the teacher) is used to train the compressed network (or the student). However, there are two very important distinctions. (1) In knowledge distillation, the parameters of the base (or larger or teacher) network are fixed and the output of the base network is used as a "soft label" to train the compressed (or smaller or student) network. In the paradigm of the adjoined network, both the base and the compressed network are trained together. The output of the base network influences the compressed network and vice-versa. (2) The parameters of the compressed network are shared across both the smaller and larger networks (Fig. \ref{fig:adjoinedNetwork}). We train the two networks using a novel time-dependent loss function called \textit{adjoined loss}. An additional benefit of training the two networks together is that the smaller network can have a regularizing effect on the larger network. In our experiments (Section \ref{section:experiments}), we see that on many datasets and for many architectures, the base network trained in the adjoined fashion has greater prediction accuracy than the standard situation when the base network was trained alone. We also provide theoretical justification for this observation in the \supp materials. The details of our design, the loss function and how it supports fast inference are discussed in Section \ref{section:adjoinedNetworks}.

	


	As discussed in the previous paragraph, in the AN training paradigm, all the parameters of the smaller network are shared across both the smaller and larger network. Our compression architecture design involves selecting and tuning a hyper-parameter $\alpha$, the size (or the number of parameters in each convolution layer) of the smaller network as compared against the larger base network. In our experiments (Section \ref{section:experiments}) with the AN paradigm, we found that choosing a value of $\alpha = 2 \textit{ or } 4$ as a global (same across all the layers of the network) constant typically worked well. To get more boost in compression performance, we propose the framework of Differentiable Adjoined Network (DAN). DAN uses techniques from Neural Architecture Search (NAS) to further optimize and choose the right value of $\alpha$ at each layer of our compressed model. The details of DAN are discussed in Section \ref{section:differentiableadjoinedNetworks}.

	Below are the main contributions of this work.
	\begin{enumerate}[wide, labelwidth=!, labelindent=0pt]
    	\item We propose a novel training paradigm based on \textit{Adjoined Networks} or AN, that can compress any CNN based neural architecture. This involves adjoined training where the original network and the smaller network are trained together. This has twin benefits of compression and regularization whereby the larger network (or teacher) transfers information and helps compress the smaller network while the smaller network helps regularize the larger teacher network.
    	
    	\item We further propose \textit{Differentiable Adjoined Networks}, or DAN, that adjointly learns some of the hyper-parameters of the smaller network including the number of filters in each layer of the smaller network.
    	
    	 \item We conducted an exhaustive experimental evaluation of our method and compared it against several state-of-the-art methods on datasets such as ImageNet \cite{russakovsky2015imagenet}, CIFAR-10 and CIFAR-100 \cite{krizhevsky2009cifar}. We consider different architectures such as ResNet-18,-50,-20,-32,-44,-56,-110 and DenseNet-121.
    	 On ImageNet, using adjoined training paradigm, we can compress ResNet-50 by $4\times$ with $2\times$ FLOPs reduction while achieving $75.1\%$ accuracy. Moreover, the base network gains $0.7\%$ in accuracy when compared against the same network trained in the standard (non-adjoined) fashion. We further increase the accuracy of the compressed model to $75.7\%$ by augmenting our approach with architecture search (DAN), clearly showing that it is \textit{better} to train the networks \textit{together}. Furthermore, we compare our approach against several state-of-the-art knowledge distillation methods on CIFAR-10 on various architectures like Resnet-20,-32,-44,-56, and -110. On each of these architectures the student trained using the adjoined method outperforms those trained using other methods (Table \ref{table:kd_variants}).
    	
    	
    	
    \end{enumerate}

	The paper is organized as follows. In Section \ref{section:relatedWork}, we discuss some of the other methods that are related to the discussions in the paper. In Section \ref{section:adjoinedNetworks}, we provide details of the architecture for adjoined networks and the loss function. In Section \ref{section:adjointLoss}, we show how training both the base and compressed network together provides compression (for the smaller network) as well as regularization (for the larger network). In Section \ref{section:differentiableadjoinedNetworks}, we combine AN with neural architecture search and introduce Differentiable Adjoined Networks (or DANs). In Section \ref{section:experiments}, we provide the details of our experimental results. 
In Section \ref{section:theory} of the appendix, we provide strong {theoretical guarantees} on the regularization behaviour of adjoined training.

\section{Related Work}
	\label{section:relatedWork}
	In this section, we discuss various techniques used to design efficient neural networks in terms of size and FLOPs. We also compare our approach to other similar approaches and ideas in the literature.
	
	\noindent \textbf{Knowledge Distillation} is the transfer of knowledge from a cumbersome model to a small model. \cite{hinton2015distilling} proposed teacher student model, where soft targets from the teacher are used to train the student model. This forces the student to generalize in the same manner as the teacher. Various knowledge transfer methods have been proposed recently. \cite{romero2015fitnets} used intermediate layer's information from teacher model to train thinner and deeper student model. \cite{peng2019correlation} proposes to use instance level correlation congruence instead of just using instance congruence between the teacher and student. \cite{ahn2019variational} tried to maximize the mutual information between teacher and student models using variational information maximization. \cite{park2019relational} aims at transferring structural knowledge from teacher to student. \cite{NEURIPS2018_6d9cb7de} argues that directly transferring a teacher's knowledge to a student is difficult due to inherent differences in structure, layers, channels, etc., therefore, they paraphrase the output of the teacher in an unsupervised manner making it easier for the student to understand. Most of these methods use a trained teacher model to train a student model. In contrast in this work, we train both the teacher and the student together from scratch. In recent work, \cite{zhang2018deep}, rather than using a teacher to train a student, they let a cohort of students train together using a distillation loss function. In this paper, we consider a teacher and a student together rather than using a pre-trained teacher. We also use a novel time-dependent loss function. Moreover, we also provide theoretical guarantees on the efficacy of our approach. We have compared our AN with various knowledge distillation methods in the experiments section.

	\noindent \textbf{Pruning} techniques aim to achieve network compression by removing parameters or weights from a network while still maintaining accuracy. These techniques can be broadly classified into two categories; unstructured and structured. Unstructured pruning methods are generic and do not take network architecture (channel, filters) into account. These methods induce sparsity based on some pre-defined criteria and often achieve a state-of-the-art reduction in the number of parameters. However, one drawback of these methods is that they are often unable to provide inference time speed-ups on commodity hardware due to their unstructured nature. Unstructured sparsity has been extensively studied in \cite{han2015deep, zhu2017prune, gale2019state, kusupati2020soft, evci2021rigging}. Structured pruning aims to address the issue of inference time speed-up by taking network architecture into account. As an example, for CNN architectures, these methods try to remove entire channels or filters, or blocks. This ensures that the reduction in the number of parameters also translates to a reduction in inference time on commodity hardware. For example, ABCPruner \cite{lin2020channel} decides the convolution filters to be removed in each layer using an artificial bee colony algorithm. \cite{lin2020hrank} prunes filters with low-rank feature maps. \cite{you2019gate} uses Taylor expansion to estimate the change in the loss function by removing a particular filter, and finally removes the filters with max change.
	The AN compression technique proposed in this paper can also be thought of as a structured pruning method where the architecture choice at the start of training fixes the convolution filters to be pruned and the amount of pruning at each layer. Another related work is of Slimmable Networks \cite{yu2018slimmable}. Here different networks (or architectures) are \textit{switched on} one at a time and trained using the standard cross-entropy loss function. By contrast, in this work, both the networks are trained together at the same time using a novel loss function (adjoined-loss). We have compared our work with Slimmable Networks in  Table \ref{table:comparison}.
	
	\noindent \textbf{Neural Architecture Search} (NAS) is a technique that automatically designs neural architecture without human intervention. The best architecture could be found by training all architectures in the given search space from scratch to convergence but this is computationally impractical. Earlier studies in NAS were based on RL \cite{zoph2017neural, tan2019mnasnet} and EA \cite{real2017largescale}, however, they required lots of computation resources. Most recent studies \cite{liu2019darts, cai2019proxylessnas, wu2019fbnet} encode architectures as a weight sharing a super-net and optimize the weights using gradient descent. A recent study Meta Pruning \cite{liu2019metapruning} searches over the number of channels in each layer. It generates weights for all candidates and then selects the architecture with the highest validation accuracy. A lot of these techniques focus on designing compact architecture from scratch. In this paper, we use architecture search to help guide the choice of architecture for compression, that is, the fraction of filters which should be removed from each layer.
	
	
	\noindent \textbf{Small architectures} - Another research direction that is orthogonal to ours is to design smaller architectures that can be deployed on edge devices, such as SqueezeNet \cite{iandola2016squeezenet},  MobileNet \cite{sandler2018mobilenetv2} and EfficientNet \cite{tan2019efficientnet}. In this paper, our focus is to compress existing architectures while ensuring inference time speedups as well as maintaining prediction accuracy.
	

	\section{Adjoined networks}
	\label{section:adjoinedNetworks}

	In our training paradigm, the original (larger) and the smaller network are trained together. The motivation for this kind of training comes from the principle that \textit{good teachers are lifelong learners}. Hence, the larger network which serves as a teacher for the smaller network should not be frozen (as in standard teacher-student architecture designs \cite{hinton2015distilling}). Rather both should learn together in a "combined learning environment", that is, adjoined networks. By learning together both the networks can be better together.
	
	We are now ready to describe our approach and discuss the design of adjoined networks. Before that, let's take a re-look at the standard convolution operator. Let $\mb x \in \mb R^{h \times w \times c_{in}}$ be the input to a convolution layer with weights $\mb W \in \mb R^{c_{out} \times k \times k \times c_{in}}$ where $c_{in}, c_{out}$ denotes the number of input and output channels, $k$ the kernel size and $h, w$ the height and width of the image. Then, the output of the convolution $\mb z$ is given by
	\begin{align*}
	&\mb z = conv(\mb x, \mb W)
	\end{align*}
	In the adjoined paradigm, a convolution layer with weight matrix $\mb W$ and a binary mask matrix $M \in \{0, 1\}^{c_{out} \times k \times k \times c_{in}}$ receives two inputs $\mb x_1$ and $\mb x_2$ of size $h\times w\times c_{in}$ and outputs two vectors $\mb z_1$ and $\mb z_2$ as defined below.
	\begin{align}
	&\mb z_1 = conv(\mb x_1,  \mb W)
	&\mb z_2 = conv(\mb x_2,  \mb W*M) \label{eqn:adjoinedNetworks}
	\end{align}
	Here $M$ is of the same shape as $\mb W$ and $*$ represents an element-wise multiplication. Note that the parameters of the matrix $M$ are fixed before training and not learned. The vector $\mb x_1$ represents an input to the original (bigger) network while the vector $\mb x_2$ is the input to the smaller, compressed network. For the first convolution layer of the network $\mb x_1 = \mb x_2$ but the two vectors are not necessarily equal for the deeper convolution layers (Fig. \ref{fig:adjoinedNetwork}). The mask matrix $M$ serves to zero-out some of the parameters of the convolution layer thereby enabling network compression. In this paper, we consider matrices $M$ of the following form.
	\begin{equation} \label{eqn:adjointMatrix}
    \begin{split}
       M := a_\alpha & = \text{ matrix such that the first $\alpha$ filters }
        \text{ are all 1 and the rest 0}
    \end{split}
    \end{equation}

	In Section \ref{section:experiments}, we run experiments with $M := a_\alpha$ for  $\alpha \in \{2, 4, 8, 16\}$. Putting this all together, we see that any CNN-based architecture can be converted and trained in an adjoined fashion by replacing the standard convolution operation by the adjoined convolution operation (Eqn. \ref{eqn:adjoinedNetworks}).  Since the first layer receives a single input (Fig. \ref{fig:adjoinedNetwork}), two copies are created which are passed to the adjoined network. The network finally gives two outputs $\mb p$ corresponding to the original (bigger or unmasked) network and $\mb q$ corresponding to the smaller (compressed) network, where each convolution operation is done using a subset of the parameters described by the mask matrix $M$ (or $M_\alpha$).  We train the network using a novel time-dependent loss function which forces $\mb p$ and $\mb q$ to be close to one another (Defn. \ref{defn:adjointLoss}).
	
	\begin{figure*}[t]
		\vspace{-30pt}
		\centering
		\subfloat{{\includegraphics[width=3.5cm, trim=10 10 8 8, clip]{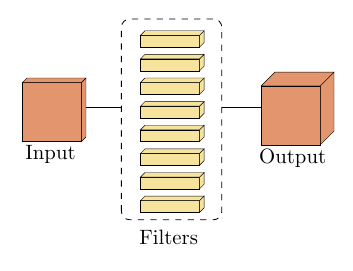} }}%
		\qquad
		\subfloat{{\includegraphics[width=4cm, trim=10 10 10 8, clip]{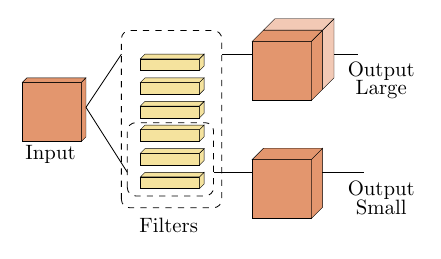} }}%
		\qquad
		\subfloat{{\includegraphics[width=4.5cm, trim=10 8 10 8, clip]{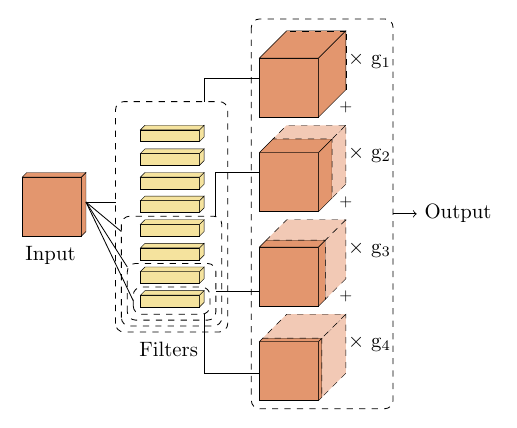} }}%
		\caption{(Left) Standard layer of CNN. (Center) Layer in Adjoined Network. (Right) Layer in DAN.}%
		\label{fig:DAN}
	\end{figure*}
	\vspace{-5pt}
	
\section{Regularization and Compression }
	\label{section:adjointLoss}
	In the previous section, we looked at the design on adjoined networks. For one input $(\mb X, \mb y) \in \mb R^{h \times w \times c_{in}} \times [0, 1]^{n_{c}}$, the network outputs two vectors $\mb p$ and $\mb q \in [0, 1]^{n_{c}}$ where $n_c$ denotes the number of classes and $c_{in}$ denotes the number of input channels (equals $3$ for RGB images).
	
	\begin{definition}[Adjoined loss]
		\label{defn:adjointLoss}
		Let $y$ be the ground-truth one-hot encoded vector and $ p$ and $q$ be output probabilities by the adjoined network. Then
		\begin{align}
		&\mc L(y, p, q) \enspace=\enspace  -y\log p \enspace+ \enspace\lambda(t) \thinspace KL(p, q) \label{eqn:adjointLoss}
		\end{align}
		where $KL(p, q) = \sum_i p_i \log \frac{p_i}{q_i}$ is the measure of  difference between two probability measures  \cite{kullback1951information}. The regularization term $\lambda: [0, 1] \rightarrow \mb R$ is a function which changes with the number of epochs during training. Here $t = \frac{\text{current epoch}}{\text{Total number of epochs}}$ equals zero at the start of training and equals one at the end.
	\end{definition}
	
	In our definition of the loss function, the first term is the standard cross-entropy loss function which trains the bigger network. To train the smaller network, we use the predictions from the bigger network as a soft ground-truth signal. We use KL-divergence to measure how far the output of the smaller network is from the bigger network. This also has a regularizing effect as it forces the network to learn from a smaller set of parameters. Note that, in our implementations, we use $KL(p, q) = \sum p_i \log \frac{p_i + \epsilon}{q_i + \epsilon}$ to avoid rounding and division by zero errors where $\epsilon = 10^{-6}$.
	
	At the start of training, $p$ is not a reliable indicator of the ground-truth labels. To compensate for this, the regularization term $\lambda$ changes with time. In our experiments, we used $\lambda(t) = \min \{4t^2, 1\}$. Thus, the contribution of the second term in the loss is zero at the beginning and steadily grows to one at $50\%$ training.


	\section{DAN: Differentiable Adjoined Networks }
	\label{section:differentiableadjoinedNetworks}
	
	In Sections \ref{section:adjoinedNetworks} and \ref{section:adjointLoss}, we described the framework of adjoined networks and the corresponding loss function. An important parameter in the design of these networks is the choice of parameter $\alpha$. Currently, the choice of $\alpha$ is global, that is, we choose the same value of $\alpha$ for all the layers of our network. However, choosing $\alpha$ independently for each layer would add more flexibility and possibly improve performance of the current framework. To solve this problem, we propose the framework of Differentiable Adjoined Networks (or DANs).
	
	Consider the following example of a convolution network with 1 layer with the following choices for $\alpha \in A =  \{1, 2, 4\}$ that outputs a vector $p_{\alpha}$. Finding the optimal network structure is equivalent to solving $\argmax_{\alpha \in A} L(p_\alpha)$ where $L$ is some loss function. For a one layer network, we can solve this problem by computing $L(p_\alpha)$ for all the different values and then computing the max. However, this becomes intractable as the number of layers increase; for a 50-layer network, the search space has size $3^{50}$.
	\begin{definition}[Gumbel-softmax (\cite{DBLP:conf/cvpr/WanDZHTXWYXCVG20})]
	    Given vector $v = [v_1, \ldots, v_n]$ and a constant $\tau$. The gumbel-softmax function is defined as $g(v) = [g_1, \ldots, g_n]$ where
	    \begin{equation}
    	g_i = \frac{\exp[(v_i+ \epsilon_i)/\tau] }{\sum_i{\exp[(v_i + \epsilon_i)/\tau]}}\label{eqn:gumbelSoftmax}
    	\end{equation}
	and $\epsilon_i \sim N(0, 1)$ is uniform random noise (also referred to as gumbel noise). Note that as $\tau \rightarrow 0$, gumbel-softmax tends to the $\argmax$ function.
	\end{definition}
	Gumbel-softmax is a ``re-parametrization trick" that can be viewed as a differentiable approximation to the $\argmax$ function. Returning back to the one-layer example, the optimization objective now becomes $\sum_{\alpha \in A}  g_\alpha L(p_\alpha)$ where $g_\alpha$ represents the gumbel weights corresponding to the particular $\alpha$. This objective is now differentiable and can be solved using standard techniques like back-propagation.
	
	With this insight, we propose the DAN architecture (Fig \ref{fig:DAN}) where the standard convolution operation is replaced by a DAN convolution operation. As before, let $\mb x \in R^{h \times w \times c_{in}}$ be the input to the DAN convolution layer with weights $ W \in  R^{c_{out} \times k \times k \times c_{in}}$ where $c_{in}, c_{out}$ denotes the number of input and output channels, $k$ the kernel size and $h, w$ the height and width of the image. Let $A = \{\alpha_1, \ldots, \alpha_m\}$ be the range of values of $\alpha$ for the layer. Then, the output $\mb z$ of the DAN convolution layer is given by
	\begin{align*}
		& z(\eta) = \sum_{i=1}^{m}   g(\eta)_i  \enspace z_i  \numberthis \label{eqn:DnasOutput}
	\end{align*}
	where $\eta = [\eta_1, \ldots, \eta_m]$ denotes the mixing weights corresponding to the different $\alpha$'s, $g$ is the gumbel-softmax function and $z_i = conv(\mb x, W *M_i)$ where $M_i$ is the mask matrix corresponding to $\alpha_i$ (as in Eqn. \ref{eqn:adjointMatrix}). Thus, each layer of the DAN convolution layer combines its outputs according to the gumbel weights. Choosing the hyper-parameter $\alpha$ now corresponds to learning the values of the parameter $\eta$ for each layer of our DAN conv network. Note that as before, our network outputs two probability vectors $\mb p$ and $\mb q$. But these vectors now also depend upon the weights vector $\eta$ at each layer. We are now ready to define our main loss function.
	
	\begin{definition}[Differentiable Adjoined loss]
		\label{defn:differentialadjointLoss}
		Let the search space be $A = \{\alpha_1, \ldots, \alpha_m\}$Let $y$ be the ground-truth one-hot encoded vector and $ p$ and $q$ be output probabilities of the adjoined network. Then
		\begin{align}
		&\mc L(y, p, q) = -y\log p + \lambda(t) (\thinspace KL(p, q) + \gamma n_f(H)) \label{eqn:adjointDanLoss}
		\end{align}
		where $KL(p, q), \lambda{(t)}$ are the same as used in Eqn. \ref{defn:adjointLoss}. $H = [\eta_1, \ldots, \eta_l]$ where $\eta_i$ is the mixing weight vector for the $i^{th}$ convolution layer. $n_f$ represents the gumbel weighted FLOPs or floating point operations for the given network. That is,
		$$n_f(H) = \sum_{\eta_i \in H} \sum_{j=1}^m g(\eta_i)_j \enspace  FLOPs(i, \alpha_j)$$
		where $flops(i, \alpha_j)$ measures the number of floating point operations at the $i^{th}$ convolution layer corresponding to the hyper-parameter $\alpha_j$. Also, note that  $\gamma$ in Eqn. \ref{eqn:adjointDanLoss} is a normalization constant.
	\end{definition}
	
	Differentiable Adjoined Loss is similar to Adjoined Loss defined in Eqn. \ref{eqn:adjointLoss}. However, the key difference is the $n_f$ term. First note that, larger architectures tend to have higher accuracies. Hence, DAN learning tends to prefer a network with low alpha (large network) against that with high alpha (small network). Thus, the $n_f$ term acts as a regularization penalty against DAN preferring large architectures. Another point to note is that for a large network say Resnet-50, the number of flops corresponding to any setting of the mixing weights can be very large. Gamma normalizes it so that all the terms in the loss function are in the same scale.

	\section{Experiments}
	\label{section:experiments}
	
	We are now ready to describe our experiments in detail. We run experiments on three different datasets. (1) \textit{ImageNet}  - an image classification dataset \cite{russakovsky2015imagenet} with 1000 classes and about $1.2M$ images . (2) \textit{CIFAR-10} - a collection of $60k$ images in 10 classes. (3) \textit{CIFAR-100} - same as CIFAR-10 but with 100 classes \cite{krizhevsky2009cifar}. For each of these datasets, we use standard data augmentation techniques such as random-resize cropping, random flipping. 
	
	We train different architectures such as ResNet-100, ResNet-50, ResNet-18, ResNet-110, ResNet-56, DenseNet-121 on all of the above datasets. On each dataset, we first train these architectures in the standard non-adjoined fashion using the cross-entropy loss function. We will refer to it by the name \textit{Standard}. Next, we train the adjoined network, obtained by replacing the standard convolution operation with the adjoined convolution operation, using the adjoined loss function. In the second step, we obtain two different networks. In this section, we refer to them by \textit{AN-X-Full $a_\alpha$} and the \textit{AN-X-Small $a_\alpha$} networks where X represents the number of layers and $a_\alpha$ denotes the mask matrix as defined in \ref{eqn:adjointMatrix}. For example, \textit{AN-50-Full $a_2$}, \textit{AN-50-Small $a_2$} represents larger and smaller networks obtained on adjoinedly training ResNet-50 with $\alpha=2$. \textit{AN-121-Full $a_4$}, \textit{AN-121-Small $a_2$} represents models obtained on adjoinedly training DenseNet-121 with $\alpha=4$. We compare the performance of the AN-X-Full $a_\alpha$ and AN-X-Small $a_\alpha$ networks against the standard network. One point to note is that we do not replace the convolutions in the stem layers but only those in the residual blocks. Since most of the weights are in the later layers, this leads to significant space and time savings while retaining competitive accuracy. DAN describes the performace of adjoined network on architectures found by Differentiable Adjoined Network. DAN-50 has the same number of blocks as ResNet-50 whereas DAN-100 has twice the number of blocks of ResNet-50. 
	
	We ran our experiments on GPU enabled machine using Pytorch.  \ifdefined\COMPLETE We have also open-sourced our implementation \footnote{The code can be found at \url{https://github.com/utkarshnath/Adjoint-Network.git}}. \else We have also uploaded the source code in the supplementary material along with a readme file.
	\fi
	Hyperparameters for the experiments are mentioned on our github page.


	 In Section \ref{subsection:exp-comparison}, we compare our compression results against other structured pruning methods. In Section \ref{subsection:exp-compression-kd}, we compare AN with various types of knowledge distillation methods. In Section \ref{subsection:exp-compression}, we describe our results for compression and performance of architectures found by DAN. In Section \ref{subsection:exp-regularization}, we show the strong regularizing effect of AN training. 
	 
	
	\vspace{-10pt}
	\subsection{Comparison against other Structured Pruning works}
	\label{subsection:exp-comparison}

	\begin{table*}[h]
	\vspace{-10pt}
		\footnotesize
		\centering
		\renewcommand{\arraystretch}{1.0}
		\begin{tabular}{|p{3.2cm}|p{1.3cm}|p{1.3cm}|p{1.3cm}|}
			\hline
			\multicolumn{4}{|c|}{{Model Compression Results}}\\
			\hline
			\multirow{1}{*}{{Method}} &
			\multirow{1}{*}{{\# Params}}
			&\multirow{1}{*}{{GFLOPs}}
			& \multirow{1}{*}{{Accuracy}} \\
			\hline
			{ABCPruner-$0.8$} {\cite{lin2020channel}} & {11.75} & {1.89}  &{73.86}\\
			{ABCPruner-$0.7$} {\cite{lin2020channel}} & {11.24} & {1.8} & {73.52} \\
			{GBN-$50$} {\cite{you2019gate}}& {11.91} & {1.9}  & {75.18} \\
			{GBN-$60$} {\cite{you2019gate}}& {17.6} & {2.43} &{76.19}   \\
			{DCP} {\cite{zhuang2018discrimination}} & {12.3} & {1.82} & {74.95}  \\
			{HRank} {\cite{lin2020hrank}}& {16.15} & {2.3} & {74.98}   \\
			{HRank} {\cite{lin2020hrank}} & {13.77} & {1.55} & {71.98}  \\
			{HRank} {\cite{lin2020hrank}} & {8.27} & {0.98} & {69.1 }   \\
			{MetaPruning} {\cite{liu2019metapruning}} & {19.1} & {2} & {75.4} \\
			{MetaPruning} {\cite{liu2019metapruning}}& {12.7} & {1} & {73.4} \\
			{Slimmable Net} {\cite{yu2018slimmable}}& {19.2} & {2.3} & {74.9}\\
			{Slimmable Net} {\cite{yu2018slimmable}} & {12.8} & {1.1} & {72.1} \\
			{AN-50-Small $a_4$ (our)} & {2.2} & {1.6}
			&{71.82} \\
	
			{DAN-50 (our)} & {3.49} & {1.7} & {73.33}\\
			
			{DAN-100 (our)} & {6.58} & {2.15} & \textbf{75.43} \\
			
			{AN-50-Small $a_2$ (our)} & {7.14} & {2.2} & {75.1} \\
			\hline
		\end{tabular}
		\captionof{table}{The table shows the performance of various structured pruning methods when trained on the ImageNet dataset. $a_\alpha$ in AN-50-Small denotes the mask matrix as defined in Eqn. \ref{eqn:adjointMatrix}.}
		\label{table:comparison}
	\end{table*}
	
	\begin{figure*}[t]

		\centering
		\subfloat{{\includegraphics[scale=0.45, trim=0 0 0 0, clip]{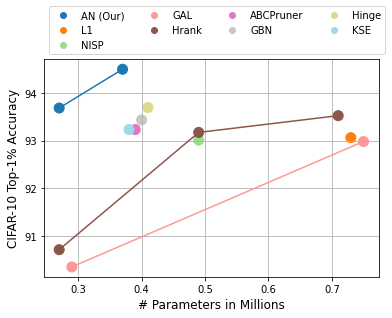} }}%
		\qquad
		\subfloat{{\includegraphics[scale=0.45, trim=0 0 0 0, clip]{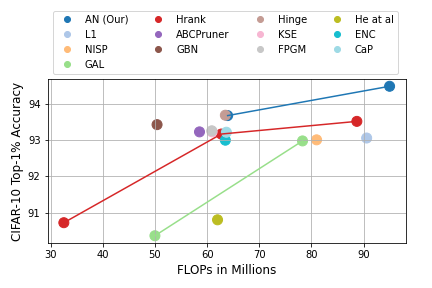} }}%
		\qquad
		\caption{Top-1\% accuracy of various pruning methods (by compressing ResNet-56 architecture) on CIFAR-10 dataset plotted against number of parameters (Left) and FLOPs (Right). Pruning methods - Gal(\cite{lin2019optimal}), Hrank(\cite{lin2020hrank}) ,He at al(\cite{lin2019optimal}), ENC(\cite{kim2019efficient}),NISP(\cite{yu2018nisp}), L1(\cite{li2017pruning}), ABC-Prunner(\cite{lin2020channel}), CaP(\cite{minnehan2019cascaded}), KSE(\cite{li2019exploiting}), FPGM(\cite{he2019filter}), GBN(\cite{you2019gate}) and Hinge(\cite{li2020group})}%
		\label{fig:comparison_cifar10}
	\end{figure*}

	Table \ref{table:comparison} and Figure \ref{fig:imagenetComparison} compare the performance of various structured compression schemes on the ImageNet dataset for the ResNet-50 architecture. Note, these methods provide inference speed-up without special hardware or software. We see that the adjoined training regime can achieve compression that is significantly better than other methods considered in the literature. In Figure \ref{fig:imagenetComparison}, models trained using our paradigm are explicitly on the left side of the graph while other methods are clustered on to the right side. Other methods obtain compression ratios in the range $2-3\times$, compared to which our method achieves up to $12\times$ compression in size. Similarly, GFLOPS for our method is amongst the highest as compared to the other state-of-the-art works, while suffering a small accuracy drop as compared against the base ResNet-50 model.
	Figure \ref{fig:comparison_cifar10} compares the performance of AN against various pruning methods on CIFAR-10 dataset for ResNet-56 architecture. Models trained using AN paradigm achieve highest accuracy with fewest number of parameters on CIFAR-10. AN exceeds the next best model (Hinge \cite{li2020group}) by 0.8\% while being smaller than 9 of the 11 models. The smallest AN model achieves accuracy similar to hinge but with 35\% fewer parameter. We see similar results for FLOPs. 

    \subsection{Comparison against other Knowledge Distillation Works}
    \label{subsection:exp-compression-kd}
	
	\begin{table*}[h]
		\scriptsize
		\centering
		\renewcommand{\arraystretch}{1.0}
		\begin{tabular}{ |p{1.3cm}|p{1.8cm}|p{1.8cm}|p{1.8cm}|p{1.8cm}|p{1.8cm}|}
			\hline
			\multicolumn{6}{|c|}{Knowledge Distillation Variants}\\
			\hline
			{Method} &{AN-110-Small} &{AN-56-Small} &{AN-44-Small} &{AN-32-Small} &{AN-20-Small}\\
			\hline
RKD	\cite{park2019relational}& 94.11	& 93.72	& 93.41	& 92.5	& 92.15 \\
VID	\cite{ahn2019variational}& 93.6	& 93.23	& 92.86	& 92.41	& 91.6 \\
FT	\cite{NEURIPS2018_6d9cb7de}& 93.76	& 93.33	& 93.17	& 92.49	& 91.2 \\
CC	\cite{peng2019correlation}& 93.6	& 93.22	& 92.95	& 92.46	& 91.5 \\
DML	\cite{zhang2018deep}& 93.5	& 93.3	& 93.2	& 92.59	& 91.35 \\
KR	\cite{chen2021reviewkd}& 94.08	& 93.85	& 93.51	& 93.45	& 92.3 \\
KDCR \cite{Guo_2020_CVPR}	& 94.26	& 93.7	& 93.4	& 92.9	& 91.85 \\
AN	(Our)& \textbf{95} &	\textbf{94.49} & \textbf{94.01} & \textbf{93.45} & \textbf{92.45}\\
            \hline

		\end{tabular}
		\caption{AN compared to various state-of-the-art KD methods on CIFAR-10. All AN-X-Small models refers to models with $\alpha=2$.}
		\label{table:kd_variants}
	\end{table*}


	In this section, we discuss the effectiveness of weight sharing and training two networks together. We compare AN against the various state-of-the-art variants of knowledge distillation. In Table \ref{table:kd_variants}, we compare accuracy (Top-1\%) of AN-X-Small against the same architecture trained using various KD variants on CIFAR-10 dataset. The corresponding pre-trained ResNet architecture was used as the teacher model for KD variants. Teacher models were trained on CIFAR-10 using standard training paradigm. We see that all models trained using AN paradigm significantly outperforms the models trained using various teacher-student paradigm showing the effectiveness of training a subset of weights together.

	\subsection{Ablation study: Compression}
	\label{subsection:exp-compression}
	
\begin{table*}[h]
		\scriptsize
		\centering
		\renewcommand{\arraystretch}{1.0}
		\begin{tabular}{|p{2.5cm}|p{1.2cm}|p{1.2cm}|p{1.2cm}|}
			\hline
			\multicolumn{4}{|c|}{Compression using AN paradigm}\\
			\hline
			\multirow{1}{*}{Network} &
			\multirow{1}{*}{\#Params} &
			\multirow{1}{*}{GFLOPs} & \multirow{1}{*}{\begin{tabular}[c]{@{}c@{}}{Accuracy}\end{tabular}} \\

			\hline
			\multicolumn{4}{|c|}{CIFAR-10}\\
			\hline
			ResNet-20           & 0.27	& 40  & 92.5\\
			AN-20-Small $a_{2}$	& 0.07	& 21  & 92.45\\
            ResNet-32		    & 0.46	& 69  & 93.1\\
            AN-32-Small $a_{2}$	& 0.13	& 35  & \textbf{93.45}\\
            ResNet-44		    & 0.65	& 97  & 93.5\\
            AN-44-Small $a_{2}$	& 0.19	& 49  & \textbf{94.01}\\
            ResNet-56		    & 0.84	& 127 & 93.9\\
            AN-56-Small $a_{2}$	& 0.24	& 63  & \textbf{94.49}\\
            ResNet-110		    & 1.72	& 253 & 94.3\\
            AN-110-Small $a_{2}$& 0.49	& 127 & \textbf{95.0}\\
			\hline
			\multicolumn{4}{|c|}{ImageNet}\\
			\hline
            ResNet-50 &   25.5 & 4774 & 76.1 \\
            AN-50-Small $a_2$  & 7.14   & 2202 &  75.1 \\
			AN-50-Small $a_4$ & 2.2   & 1619 &  71.84 \\
			DAN-50 & 3.49 & 1745 & 73.33 \\
			ResNet-100 &  46.99 & 8473 & 77.3 \\
			AN-100-Small $a_4$  & 3.86    & 2681 &  74.51 \\
			DAN-100 &  6.58 & 2153 &\textbf{75.43} \\
			
			\hline
			
		\end{tabular}
		\caption{$a_\alpha$ denote the masking matrix (defined in Eqn. \ref{eqn:adjointMatrix}).}
		\label{table:compression}
	\end{table*}
	
	In this section, we evaluate the performance of models compressed by Adjoined training paradigm. Table \ref{table:compression} compares the performance (top 1\% accuracy) of the models compressed using AN against the performance of standard network. For AN, we use the $a_\alpha$ as the masking matrix (defined in Eqn. \ref{eqn:adjointMatrix}). The mask is such that the last $(1-\frac{1}{\alpha})$ filters are zero. Hence, these can be pruned away to support fast inference. For CIFAR-10, 4 out of 5 models compressed using AN paradigm exceed it's base architecture by 0.5\%-0.8\%. These models achieves 3.5-4$\times$ reduction in parameters and 2$\times$ reduction in FLOPs.
	
	We also observe that ResNet-50 is a bigger network and can be compressed more. Also, different datasets can be compressed by different amounts. For example, on CIFAR-100 dataset, the network can be compressed by factors $\sim 35\times$ while for other datasets it ranges from $2\times$ to $12\times$. DAN is able to search compressed architecture with minimum loss in accuracy as compared to base architecture. For ImageNet, DAN architectures were searched on Imagewoof (a proxy dataset with 10 different dog breeds from ImageNet \cite{howard2019imagenette,shleifer2019using}).
	$\gamma$ as defined in Defn. \ref{defn:differentialadjointLoss} is $e^{-13}$, $e^{-19}$ for DAN-50 and DAN-100 respectively. During architecture search, temperature $\tau$ in gumbel softmax was initialized to 15 and exponentially annealed by $e^{-0.045}$ every epoch.

	\subsection{Ablation study: Regularization}
	\label{subsection:exp-regularization}	
	
	\begin{table*}[h]
		\scriptsize
		\centering
		\renewcommand{\arraystretch}{1}
            \begin{tabular}{ |p{1.9cm}|p{0.3cm}|p{1cm}|p{1cm}|  }
			\hline
			\multicolumn{4}{|c|}{{AN-Full vs Standard}}\\
			\hline
			\multirow{1}{*}{{Network}} &\multirow{1}{*}{{$\alpha$}}  & \multirow{1}{*}{{AN-Full}} & \multirow{1}{*}{{Standard}}\\
			\hline
			\multicolumn{4}{|c|}{{CIFAR-10}}\\
			\hline
			ResNet-20 & $a_{2}$ & \textbf{93.51}	& 92.5\\
            ResNet-32	& $a_{2}$ & \textbf{94.39}	& 93.1\\
            ResNet-44	& $a_{2}$ & \textbf{94.64}	& 93.5\\
            ResNet-56	& $a_{2}$ & \textbf{95.01}	& 93.9\\
            ResNet-110	& $a_{2}$ & \textbf{95.4}	& 94.3\\
            \hline
			\multicolumn{4}{|c|}{{CIFAR-100}}\\
			\hline
			{ResNet-50} & {$a_{8}$} & {77.36} & {76.8} \\
			{ResNet-18} &  {$a_{2}$} &  {74.8} & {74.3}
			\\
			{DenseNet-121} & {$a_{4}$} &  \textbf{{80.8}} & {79.0}\\
			\hline
			\multicolumn{4}{|c|}{ImageNet}\\
			\hline
			{ResNet-50}  & {$a_{2}$} &  \textbf{{76.87}} & {76.1} \\
			{ResNet-50} & {$a_{4}$} &  {75.84} & {76.1} \\
			
			\hline
			
			
		\end{tabular}
		
		\caption{$a_\alpha$ denote the masking matrix (defined in Eqn. \ref{eqn:adjointMatrix}).}
		\label{table:regularization}
\end{table*}
	
	In this section, we study the regularization effect of Adjoined training paradigm on AN-Full network. Table \ref{table:regularization} compares the performance of the base network trained in adjoined fashion (AN-Full) to the same network trained in Standard fashion. We see a consistent trend that the network trained adjoinedly outperforms the same network trained in the standard way. We see maximum gains on CIFAR-100, exceeding accuracy by as much as $1.8\%$. Even on ImageNet, we see a gain of about $0.77\%$.

	\section{Conclusion}
	In this work, we introduced the paradigm of Adjoined Network training where both the larger teacher (or base) network and the smaller student network are trained together. We showed how this approach to training neural networks can allow us to reduce the number of parameters of large networks like ResNet-50  by $12\times$, (even going up to $35\times$ on some datasets) without significant loss in classification accuracy with $2$-$3\times$ reduction in the number of FLOPs. We showed (both theoretically and experimentally) that adjoining a large and a small network together has a regularizing effect on the larger network. We also introduced DAN, a search strategy that automatically selects the best architecture for the smaller student network. Augmenting adjoined training with DAN, the smaller network achieves accuracy that is close to that of the base teacher network.
		

\bibliography{ref}

\appendix
	\section{Regularization theory}
	\label{section:theory}
	\begin{theorem}
		\label{thm:regularization}
		Given a deep neural network $\mc A$ which consists of only convolution and linear layers. Let the network use one of $f(x) = \min\{x, 0\}$ (relu) or $f(x) = x$ (linear) as the activation function. Let the network be trained using the adjoined loss function as defined in Eqn. 3. Let $\mb X$ be the set of parameters of the network $\mc A$ which is shared across both the smaller and bigger networks.  Let $\mb Y$ be the set of parameters of the bigger network not shared with the smaller network. Let $\mb p$ be the output of the larger network and let $\mb q$ be the output of the smaller network where $\mb p_i$, $\mb q_i$ represents their $\mb i^{th}$ component. Then, the adjoined loss function induces a data-dependent regularizer with the following properties.
		\begin{itemize}
			\item For all $x \in X$, the induced $L_2$ penalty is given by $\sum_i \mb p_i \big(\log' \mb p_i - \log' \mb q_i\big)^2$
			\item For all $y \in Y$, the induced $L_2$ penalty is given by $\sum_i \mb p_i \big(\log' \mb p_i \big)^2$
		\end{itemize}
	\end{theorem}
	\begin{proof}
		We are interested in analyzing the regularizing behavior of the following loss function. $-y \log p + KL(p, q)$ $y$ is the ground truth label, $p$ is the output probability vector of the bigger network and $q$ is the output probability vector of the smaller network. Recall that the parameters of smaller network are shared across both. We will look at the second order taylor expansion for the kl-divergence term. This will give us insights into regularization behavior of the loss function.
		
		Let $x$ be a parameter which is common across both the networks and $y$ be a parameter in the bigger network but not the smaller one.
		\begin{flalign*}
		&D(x) = \sum_i p_i(x) \big( \log p_i(x) - \log q_i(x)\big) \text{ and } D(y) = \sum_i p_i(y) \big( \log p_i(y) - \log q_i\big) &
		\end{flalign*}
		For the parameter $y$, $q_i$ is a constant. Now, computing the first order derivative, we get that
		\begin{flalign*}
		&D'(x) = \sum_i p_i'(x) \big( \log p_i(x) - \log q_i(x)\big) + p_i'(x) - \frac{q_i'(x) p_i(x)}{q_i(x)}&\\
		&D'(y) = \sum_i p_i'(y) \big( \log p_i(y) - \log q_i\big) + p_i'(y) &
		\end{flalign*}
		
		\noindent Now, computing the second derivative for both the types of parameters, we get that
		\begin{flalign*}
		&D''(x) = \sum_i p_i''(x) \big( \log p_i(x) - \log q_i(x)\big) +  p_i'(x)\Bigg(\frac{p_i'(x)}{p_i(x)} - \frac{q_i'(x)}{q_i(x)}\Bigg) + p_i''(x) &\\
		&- \frac{q_i(x) q_i'(x) p_i'(x) + q_i(x) q_i''(x) p_i(x) - q_i'(x) q_i'(x) p_i(x)}{q_i^2(x)}&\\
		&D''(y) = \sum_i p_i''(y) \big( \log p_i(y) - \log q_i\big) +  \frac{p_i'(y) p_i'(y)}{p_i(y)} + p_i''(y)&
		\end{flalign*}

		\begin{flalign*}
		&D''(x) = \sum_i  \frac{p_i'(x) p_i'(x)}{p_i(x)} - \frac{2p_i'(x) q_i'(x)}{q_i(x)} + \frac{ q_i'(x) q_i'(x) p_i(x)}{q_i^2(x)}&
		\end{flalign*}

		\begin{flalign}
		&= \sum_i p_i(x)
		\Bigg(\frac{p_i'(x)}{p_i(x)} - \frac{q_i'(x)}{q_i(x)}\Bigg)^2= \sum_i p_i (\log' p_i - \log ' q_i)^2 \label{eqn:dppx}&
		\end{flalign}
		
		\noindent Similarly, for the parameters only in the bigger network, we get that
		\begin{flalign}
		&D''(y) =  \sum_i \frac{p_i'(y) p_i'(y)}{p_i(y)} = \sum_i p_i (\log ' p_i)^2&\label{eqn:dppy}
		\end{flalign}
		
		Note that $y$ represents the over-parameterized weights of the model. The equations above show that the regularization imposed by the KL-divergence term on these parameters is such that if these parameters change a lot (on the log scale) then the penalty imposed on such parameters is more. Thus, the kl-divergence term encourages such parameters not to change by a lot.
	\end{proof}

\end{document}